\documentclass[11pt,twocolumn]{article}

\usepackage{amssymb}
\usepackage{graphicx}
\usepackage{xcolor}
\usepackage{svg}
\usepackage{hyperref}

\usepackage{moreverb}
\def\mpar#1{\marginpar{\raggedright\scriptsize\sf #1}}

\title{RoboCup@Home Education 2020  Best Performance: 
RoboBreizh, a modular approach}

\author{
A. Dizet $^{1,2}$
\and
C. Le Bono $^1$
\and
A. Legeleux $^1$
\and
M. Neau $^1$
\and
C. Buche $^3$ 
}

\date{
$^1$ Lab-STICC, CNRS UMR 6285, France \\
$^2$ CERVVAL, France \\
$^3$ CROSSING, CNRS IRL 2010, Australia\\
\texttt{cedric.buche@cnrs.fr}
}

\begin{document}

\maketitle

\begin{abstract} 
Every year, the Robocup@Home competition challenges teams and robots' abilities. In 2020, the RoboCup@Home Education challenge was organized online, altering the usual competition rules.
In this paper, we present the latest developments that lead the RoboBreizh team to win the contest. These developments include several modules linked to each other allowing the Pepper robot to understand, act and adapt itself to a local environment. 
Up-to-date available technologies have been used for navigation and dialogue.
First contribution includes combining object detection and pose estimation techniques to detect user's intention.
Second contribution involves using Learning by Demonstrations to easily learn 
new movements that improve the Pepper robot's skills.
This proposal won the best performance award of the 2020 RoboCup@Home Education challenge.
\end{abstract}


\section{Introduction}

The \textit{RoboCup@Home} is an annual international competition that aims to develop service robot technology for personal domestic applications. Supplementarily, the \textit{Robocup@Home Education} challenge includes simplified rules to promote educational efforts.
Both competitions introduce a set of robot tasks and capabilities to achieve the desired goals. This includes human and object recognition, safe navigation, dialogue, and object manipulation. 

This paper will describe RoboBreizh's research proposal which resulted in obtaining the Best Performance award. 
Robot's perception is handled using state-of-art algorithms that detects people and objects. We use a client server architectural pattern in which the server node is physically separated from the client.  
Besides, the position of detected objects is computed using Pepper's depth sensor camera and the 
current position of the robot. Navigation is performed with a laser-based approach allowing the Pepper to safely operate in a known environment. The robot interacts with human using dialogue, its LEDS and tablet. Pepper moves to create a natural interaction and pick up objects.


In addition  to provide a comprehensive solution that handles all the tasks of the competition, 
this paper will present original contributions in twofold.
 First, object detection and pose estimation are combined to detect persons in need of assistance. Second, Gaussian mixture models (GMM) and Gaussian Mixture Regression (GMR) are used to easily generate movements for manipulation and interaction purposes.
Regarding the detection part, our solution proposes detecting the robot's surroundings and person's pose. Combining both information in real time allows to deduce whether a person requires the robot's assistance.
Regarding the robot's movement, our solution proposes to manipulate the robot to teach new movement.
This technique enables non-expert users to easily teach movements to the robot by demonstrations. We used existing Learning by Demonstrations algorithms.
This programming technique is mainly used with robotic arms which have more degrees of freedom, in this paper we propose an adaptation for the Pepper robot.

This paper is organized as follows. The first section  discusses the architecture and environment of the system we proposed. Modules using available technologies are presented (manager, navigation and dialogue). Furthermore, the contribution of this research work related to perception and movement learning are described. Finally, our implementation and results during the 2020 RoboCup@Home Education Online Challenge are introduced. 
 

\section{Architecture and modules based on available technologies} 

In this section, we present developments made to obtain expected behaviors of the robot during the competition. More precisely, we present the parts which are not innovative from a scientific point of view, but which are of interest regarding the choices and their respective implementations.




\subsection{Architecture}
\begin{figure}[thpb]
    \centering
    \includegraphics[width=.85\linewidth]{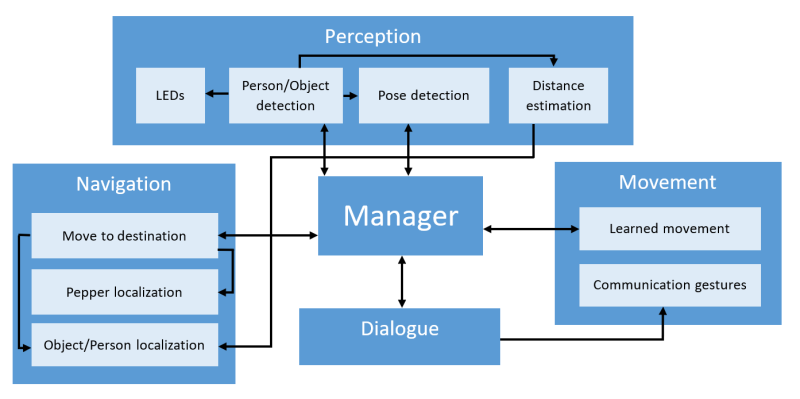}
    \caption{RoboBreizh's Architecture 
    }
    \label{architecture}
\end{figure}

Our architecture is built upon five ROS 
\cite{quigley2009ros}
modules: Manager, Perception, Navigation, Movement and Dialogue. 
Because of the limited resources of the Pepper robot, ROS and its modules run on external computers. 
As highlighted in Figure \ref{architecture}, our architecture uses ROS for its flexibility and the ability to easily segment our code into modules as follows:
\begin{itemize}
    \item  Manager:    Custom
    \item  Perception: OpenPose / Mask-RCNN
    \item  Navigation: ROS Navigation Stack
    \item  Movement:   GMM / GMR 
    \item  Dialogue:   NAOqi Python SDK 
 \end{itemize}
Next sections describe each module and their components.

\subsection{Manager module}

The manager is a high-level structure. 
Modules' output are stored in the Manager as object instances and computed results are made available on demand.
In contrast to the state machines \cite{lausen2003model}, the Manager is a custom rule-based system executing tasks in the required order and scheduling Pepper's behavior. It also handles task priority. This architecture was designed to easily add new abilities. During the competition, the Manager allowed  to quickly create four different scenarios. Shortest jobs are prioritized over time taking jobs to get rid of computational delays.


\subsection{Navigation module} 
\label{mapping}

\begin{figure}[thpb]
    \centering
    \includegraphics[scale=0.15]{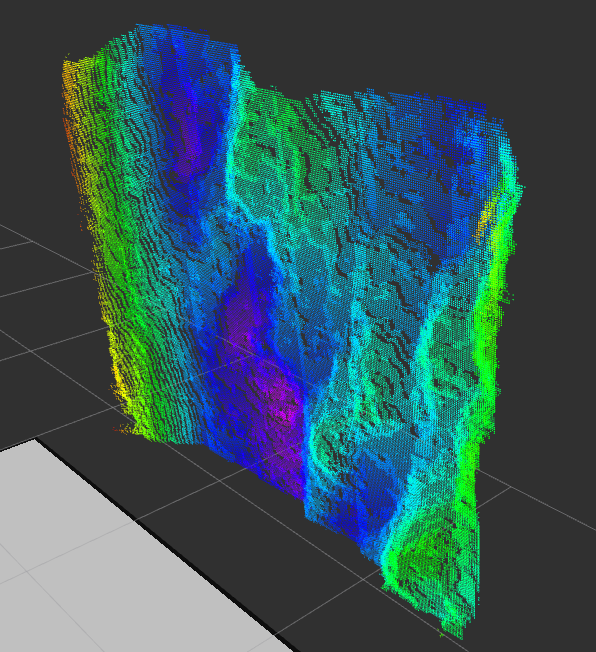}
    \caption{Pepper looking to a flat wall - Rviz visualization}
    \label{lens}
\end{figure}

As far as navigation with ROS is concerned, the two main approaches are: Visual-SLAM \cite{octomap,rtabmap} and Laser-based (e.g. LiDAR) SLAM \cite{setting_nav}. The first one computes visual information from 2D monocular cameras and 3D Depth sensors to build a 3D map.
In \cite{depth_camera}, experiments have been conducted on Pepper's depth camera (the Asus Xtion) that lead to a Root Mean Square Error (RMSE) of 20.36mm for a 1m distance and 79.15mm for 3m distance. This error is found to stem from the robot's lenses \cite{depth_camera}.
Other issues such as depth shadow or monocular depth estimation are also covered in this paper.
In our case, we experienced those issues by qualitative observations as shown in Figure \ref{lens}. Although there are various calibration methods for this sensor, \cite{calibration,depth_camera}, they are complex and can not completely eliminate the error.
Concerning the Laser-Based SLAM approach, Pepper's lasers are located in front, left and right of the robot with a VFOV of 40° and HFOV of 60° (Vertical/Horizontal Field of View). Unfortunately, those sensors are represented with only 6 beams (separated by 10° each) through ROS. This means that Pepper can only detect a maximum of 18 obstacle points at once. This is insufficient for safe navigation.
Therefore, we chose to fuzz Pepper lasers and depth camera data in a Laser-based SLAM approach. This approach is implemented in the ROS Navigation Stack \cite{setting_nav}. Data from the depth camera are first converted to point cloud and then to laser data using the depthimage\_to\_laserscan ROS package. Once a valid map is obtained, it can be used for navigation with the AMCL \cite{amcl} and the move\_base \cite{move_base} ROS nodes.


\label{nav_discuss}
We also have chosen not to implement the navigation module on board for the fact that ROS is currently not supported by the robot system. As such, lag between Pepper and ROS running on the external computer has to be taken into consideration. Notably, this had an impact on odometry as it did not match well with the current position of the robot, in particular when making fast moves or turning around. As described by \cite{nav_tuning}, other parameters such as bad laser messages or errors in dynamic reconfiguration of the path could also lead to failures.






\subsection{Dialogue module} 



Our approach for the dialog module with Pepper relies on the in-built NAOqi System. It provides a basic Automatic Speech Recognition (ASR) engine (Nuance Vocoon) and a simple rule-based system for Natural Language Processing (NLP) mostly based on key words detection. As highlighted in \cite{de2018towards}, the precision of the ASR is only about 69\% on the baseline vocabulary. However, in the context of simple action-command sentences, we believe that this is sufficient to understand the main words related to the command. Although this approach is not as efficient as Google Speech-To-Text ASR or NLTK \cite{nltk} for NLP, it has the advantage of being easy-to-use and configurable in the robot.
Text-to-Speech is also performed by the NAOqi system, as well as automatic gestures during dialog.

\section{Contribution} 
This section focuses on innovative solutions proposed by the RoboBreizh team.

\subsection{Perception module} 
\label{section_detection}

When it comes to the requirements of the competition, the ability of a robot to interact with humans and objects is crucial. To do so, it is required to detect them first. In addition to simple identification, distance estimation is also important. Moreover, beyond human detection, it was also deemed interesting enough to detect when a person raises its hand as we used it to call the robot.



Nowadays, computer vision is one of the most important part of robotics projects and is typically handled by convolutional neural network approaches \cite{Girshick_2014_CVPR}. The use of YOLO  \cite{Redmon_2016_CVPR} has become a standard in this field, especially for object and person detection. Although YOLO is the most prevalent option, other solutions are also available. One of them is Mask-RCNN \cite{DBLP:journals/corr/HeGDG17}, it provides a mask generation of objects detected in addition to the bounding boxes. Beyond object and people detection, performing pose estimation was deemed necessary to detect people's movement (e.g. waving hands). An efficient tool in this domain is OpenPose \cite{OpenPose}, a real time multi-person system which can detect up to 135 different kinds of body keypoints. The contribution here is to combine in real time information from object detection and information from keyframes. Waving hands are detected by checking whether a person's arm is located above their elbows or shoulders. 

\begin{figure}[thpb]
    \centering
    \includegraphics[width=.75\linewidth]{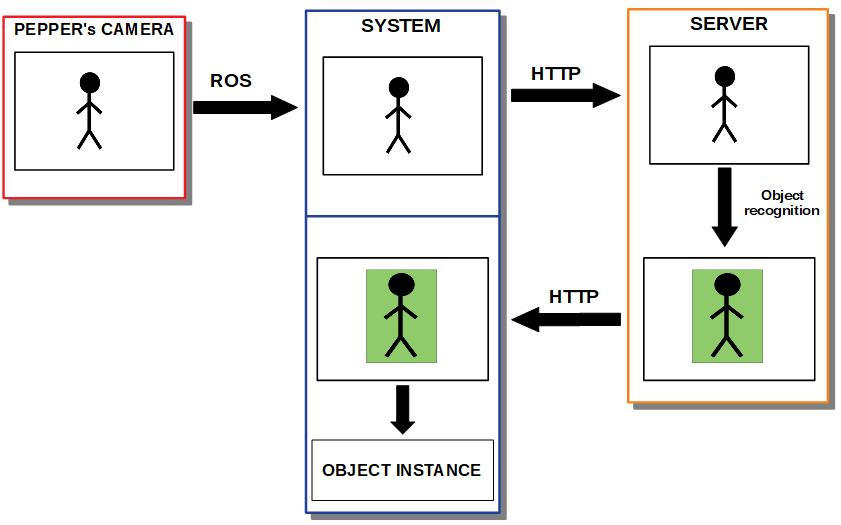}
    \caption{System server link}
    \label{server}
\end{figure}

A state-of-the-art computer vision approaches cannot be put on board a Pepper robot. A solution is to use an external computer with a dedicated GPU to handle this constraint. The Perception module is defined as a pipeline (Figure \ref{server}). The server and manager communicate using HTTP requests as it was deemed simpler than using ROS when dealing with multiple computers.
Using ROS, Pepper's camera images are retrieved and are sent to a server. This last performs object detection and extracts relevant data. From those data, high-level features such as object position are computed. This kind of architecture allows us to request the detection of specific objects (type and number, e.g. retrieving the position of at least 2 persons) with the associated features.

\begin{figure}[thpb]
    \centering
    \includegraphics[scale=0.5]{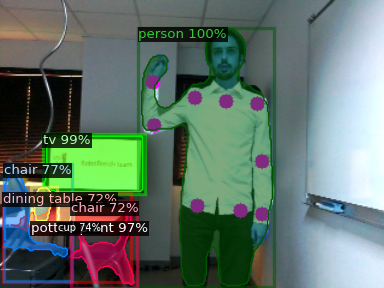}
    \caption{Human detection and object detection using Mask R-CNN}
    \label{figure_mask}
\end{figure}

The images obtained from Pepper's cameras are sent to the server and fed to Mask-RCNN \cite{DBLP:journals/corr/HeGDG17} and OpenPose. Mask R-CNN was chosen over other state-of-the-art object detection algorithms due to its ability to detect object with pixel-level precision. Compared to other solutions, this minimizes the noise added by the background and reduces the risk of inaccurately localizing the object. This level of precision is required when measuring the distance between Pepper and the object (see later). The Mask-RCNN neural network uses the default weights trained on the COCO dataset \cite{lin2014microsoft}. It can detect 80 kind of widely used objects. The combination of Mask-RCNN and OpenPose takes about 400ms on a computer with an Nvidia GTX 1060 GPU.
By using the positions of the chairs and persons in an image and how they overlap, it is possible to determine whether the chairs are available or taken. Additionally, OpenPose is used to extract the positions of all the hands in an image and thus whether someone is waving. The server then returns the positions of all the objects, persons, and waving hands detected in an image. Those positions can be used to measure the objects' distances to the robot. Our Perception module were built to work in simulation as well as in real-life as shown in Figure \ref{figure_mask}. Additionally, gender and age estimation is performed using models and weights proposed by \cite{10.1145/2818346.2830587}.

\label{section_world_representation}
As an interaction robot, Pepper needs to know the static (e.g. walls and fixed objects) and dynamic (e.g. people and non-fixed objects) parts of its environment. The static environment is modeled using a mapping tool (Section \ref{mapping}). In the 
2-Dimensional mapped environment, the position of Pepper and the objects is modelled by x and y coordinate points 
in a Cartesian coordinate system. The distance between an object and Pepper is obtained by combining the object detection (Section \ref{section_detection}) and Pepper's depth camera values. This distance is the average depth of the object's mask depth values (Figure \ref{figure_mask}). Based on the distance between an object and Pepper, the position of each detected object is then computed using trivial mathematical equations. 
Pepper's position corresponds to the central point of the robot. Those position are published in real time. The current application can be set up for any object known through supervised learning.





YOLO was originally chosen for the detection of objects and persons. This came with several drawbacks. Since YOLO implementations return a rectangular bounding box for the detection of each object, the background occupies most of the box. When measuring the distance between the object and Pepper, the object has a high chance of being incorrectly localized. Several solutions were tested to solve that issue. First, measuring the distance between Pepper and the center of the bounding box only. This solution did not work when the center of the object was part of the background, which sometimes happened with chairs. The second solution was to select the average distance between Pepper and a few selected points in the bounding box. Again, this provided unsatisfying results when many of the points where part of the background. Thus, Mask R-CNN was chosen instead of YOLO. It had the advantage of solving the background problem. However, this came at a computing cost as Mask R-CNN is slower than YOLO.

Additionally, the Pepper's LEDs are used to provide feedback to the users. For instance, the color of Pepper's eyes changes when the robot has detected that someone is waving their hands and requires assistance. This way, the users can get real-time feedback about what the robot's information.


\mpar{
}

\subsection{Movement module} 

\begin{figure}[htpb]
    \centering
    \includegraphics[width=.6\linewidth]{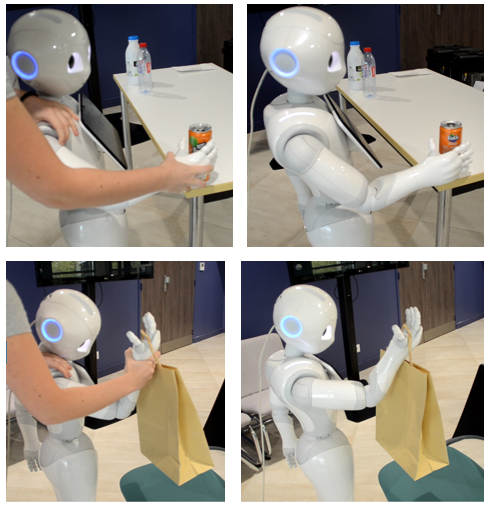}
    \caption{
    Learning by Demonstrations, movements "Take a soda" and "Take a bag". The left column displays the demonstrations and the right one, the learned movement.}
    \label{Movement}
\end{figure}

Robots move their bodies to interact with their environment and with humans. In the Robocup@Home Education, robot needs to know how to perform specific movements (e.g. take objects, point an empty seat, etc.). Learning by Demonstrations is one of the easiest ways to teach a movement to a robot as shown in Figure \ref{Movement}. We use kinesthetic demonstrations (i.e. human moves the robot's arms). With multiple demonstrations, the robot can generalize the movement. The learning is done at the trajectory level. With this contribution, no code is necessary to define gestures. 

MoveIt! \cite{trends} is an easy way to plan robotics movements, but it is not easy to create a movement compared to kinesthetic demonstrations (real world vs 2D interface). MoveIt! also depends on inverse kinematics solvers which may depend on the number of degrees of freedom (DOF) of the robot. The Pepper robot arm has 5 DOF which limits its movement in the space and reduces the chance of finding a solution with the inverse kinematics solver. Therefore, we choose to control and learn the movement in joint mode. One common approach for the learning is the Dynamic Movement Primitives (DMP) \cite{park2008movement}. DMP algorithm learns a non-linear differential equation with a single demonstration. \cite{park2008movement} develop this model to adapt the movement with different start and end position as well as the obstacle avoidance. Hidden Markov Model (HMM) allows to learn the time and space constraints \cite{calinon2010learning,billard2006discriminative}. This model uses multiple demonstrations for the learning. The start and end position can be changed, and the movement will follow the learning constraints. Gaussian Mixture Model (GMM) in addition to Gaussian Mixture Regression (GMR) learns a movement with multiple demonstrations \cite{Calinon09book}. GMM generalizes a movement with Gaussians. GMR generates the learning trajectory. \cite{kyrarini2019robot} modified the GMM/GMR to adapt the movement to environmental changes (object position and obstacle avoidance). Multiple demonstrations, different start and end position, obstacle avoidance impose to choose the modified GMM/GMR. For this scenario, a simple GMM/GMR is implemented.

\begin{figure}[htpb]
    \centering
    \includegraphics[width=.9\linewidth]{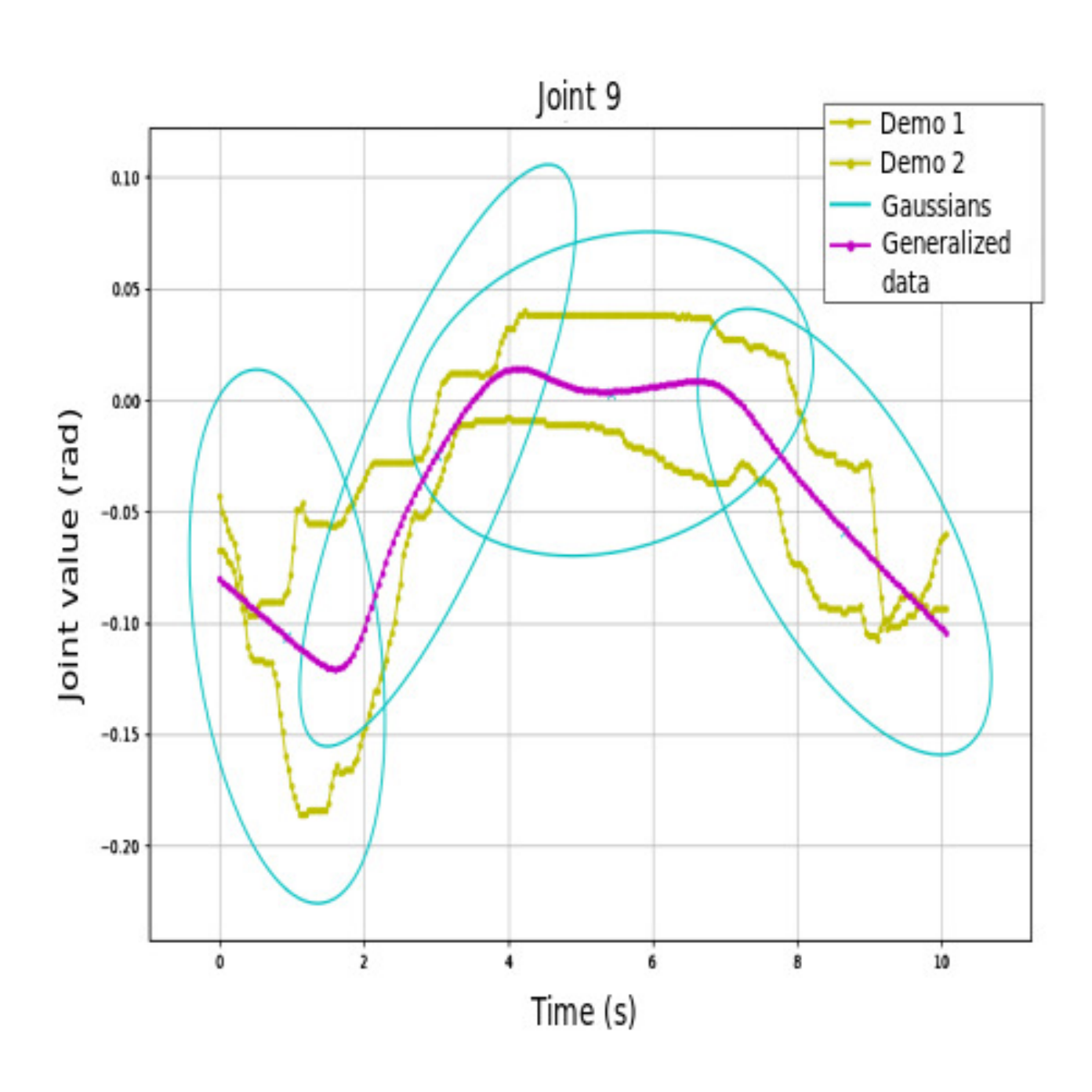}
    \caption{
   Learned movement "Point a seat" of the joint 9 with the GMM/GMR. The two demonstrations are in green, the learned movement is in pink. The four Gaussians are displayed in blue.}
    \label{Movementb}
\end{figure}

The movement module was developed under some constraints. Pepper can only pick up light objects because of its hands. The Pepper robot has limited movements due to the robot's reachable workspace. The module was developed to simplify movement learning. The learning movement module is composed of two parts: the learning phase and the movement phase. In the learning phase, the user can make multiple demonstrations to the robot for a single movement. The system automatically saves the movement (all different positions) and the user does not need to define way points of the movement. The demonstration is labelled by the user. Each demonstration is processed to remove the beginning and the end when nothing happens to all joints of the robot. The demonstrations are aligned temporally with the first demonstration of the corresponding movement. The user can start the learning process which starts GMM/GMR with previous demonstrations. The initialization of the means is done with K-means algorithm and the selection of the number of Gaussians with the BIC score. The learned movement is saved in a file. The speed of the learned movement can be set. Figure \ref{Movementb} shows the generalization of the movement "Point a seat" of the joint 9 (corresponding to the right shoulder roll) with two demonstrations. 
The user interacts with the learning phase with an interface available in French and English. This module can run a learned movement by giving the corresponding name.
The learning phase was programmed to run with various robots in simulation and in the real world. The movement phase reads the file which contains the learned movement and generate it accordingly. This phase is packed in a module used in the presented architecture (Figure \ref{architecture}). 
The Movement module works in real-time because the learning was done previously. The model works with a single demonstration or with multiple ones.

\section{Application} 
\label{Application}

The final stage of the RoboCup@Home Education Online Challenge consists an open scenario where the robot interacts with a non-expert user in a natural way. We created a scenario where the robot is a butler in the fictional RoboBreizh Hotel. Operators are a guest and the hotel's receptionist. The scenario is composed of five stages (Table \ref{Scenario}): the reception, the seat, the dialog with the director, the taxi, and the bag. 
In addition to these stages, the beginning and the end of the scenario includes a Pepper dialog for the public and the jury. The main challenges in this scenario are the navigation in a known environment with new obstacles (the guest), the perception of the environment and the physical features of a person, the movement of its arms with and without objects.

The scenario works with all modules and was presented in the Final of the RoboCup@Home Education Challenge. The Perception module works in real time and provides information about the guest. The Navigation module needs a localization in the map at each time to be able to navigate everywhere. The Dialogue module allows Pepper to dialog in a natural way with anyone. The Movement module allows Pepper to interact with different objects and perform specific movements during the conversation. By submitting this scenario in the competition, we won the Best Performance Award for Standard Platform League - Open Category. A video of our scenario is available at (\url{https://www.youtube.com/watch?v=m4rjcvv7Qa8\&t=316s}). 

\begin{table*}
\begin{tabular}{ p{2cm}  p{10cm} }
\hline

[Reception] &
Pepper waits until a client arrives, detects several physical features, moves towards the person, and asks a few questions. The discussion is altered depending on the person's detected gender. Pepper takes the reservation and saves the detected features for later. The challenges here are having a natural conversation and detecting robust features about the person. \\
\hline

[Seat] &
Pepper proposes the guest to have a seat. It turns on itself until an empty seat is in sight. Pepper then moves towards the seat, points it using a learned movement and proposes it to the guest. Pepper then asks if the client wants a drink and records the answer. This stage highlights the robot's navigation skills and its ability to find an object in its environment. \\
\hline

[Dialog] &
Pepper navigates to the receptionist's office and provides the following information: gender and name of the guest, its age. Pepper then goes back to its starting position. (It was possible to welcome other guests by restarting the reception stage. The final was limited in time, so we choose to have a single guest.) This stage shows the robot's ability to detect features and repeat them later in a different context.\\
\hline

[Taxi] &
Pepper waits for someone to wave his hand. After the detection, the robot moves toward the person and offers its services. If the guest asks for a taxi, Pepper moves its arm as if it has a phone and makes a call. Depending on the answer of the taxi station, Pepper informs the guest about the arrival of a taxi. The main challenges of this stage are the detection of a person moving his/her hand, the navigation in front of him/her and the movement of the robot's arm to make a call.\\
\hline

[Bag] &
Pepper offers its services again. If the guest asks Pepper to carry his bag. Pepper attempts to detect a bag and moves its arm to take it. If the bag is not correctly taken, the guest can help the robot. Pepper keeps his arm up and proceeds to navigate to its original position. This is the end of the scenario. This stage's challenges are the detection of the bag, the navigation in front of it, the movement to take the bag and the navigation while maintaining the bag in the hand of the robot.\\
\hline
\end{tabular}    
\caption{Five stages of our scenario}

\label{Scenario}
\end{table*}


\section{Conclusion} 

This paper describes the RoboBreizh's team approach to the RoboCup@Home Education challenge. A ROS architecture was developed to handle the competition's tasks using a Pepper robot. Notably, our robot can move to a destination, point to an empty seat, take a bag in hand, compute a person's position, talk with someone, and detect a waving hand. Our proposed architecture is also flexible and can be easily implemented on other robots. 
Our contributions are twofold. As far as vision is concerned, pose estimation and object detection are combined to extract additional information in images. Regarding movement, Learning by Demonstrations was implemented to enhance Pepper's movement skills and easily add new ones. RoboBreizh's code is open source and available on request.

Futures works will make it possible to have the code embedded. 
Concerning vision, future works include exploring newer state-of-the-art algorithms or less computationally expensive ones.   
Regarding the Movement module, movements cannot be adapted to the environment because the model used is a simple GMM/GMR. The modified GMM/GMR can overcome this constraint and add an obstacle avoidance skill. Future work will make a link with the perception module to adapt the movement in real-time.
In the future, the team will develop additional features to increase Pepper usability in person-friendly environments. In addition, we are working on a simulated environment to test all  modules before implementation 
on an actual Pepper \cite{busy2019qibullet}. This way, we could achieve more robust features while limiting the risks on our robots. 
Thanks to ROS, the same implementation can be used on the simulator and the real robot. 
Our next step will be to attempt the 2021 RoboCup@Home edition.



\section*{Acknowledgment}

This article benefited from the support of Prog4Yu ANR-18-CE10-0008, the City of Brest, CERVVAL company, the AFRAN and the Brittany Region.


\newpage
\bibliographystyle{plain}
\bibliography{ms} 
\end{document}